\documentclass{article}
\usepackage{ijcai22}

\usepackage{times}

\usepackage{soul}
\usepackage{url}
\usepackage[hidelinks]{hyperref}
\usepackage[utf8]{inputenc}
\usepackage[small]{caption}
\usepackage{graphicx}
\usepackage{amsmath}
\usepackage{booktabs}
\usepackage{floatrow}
\usepackage{algorithm}
\usepackage{algorithmic}
 \usepackage{booktabs}
\title{Deep Embedded Multi-View Clustering via Jointly Learning \\Latent Representations and Graphs}

\author{
Zongmo Huang$^1$\and
Yazhou Ren$^{1,2}$\footnote{Corresponding Author}\and
Xiaorong Pu$^{1}$\And
Lifang He$^3$\\
\affiliations
$^1$School of Computer Science and Engineering\\
University of Electronic Science and Technology of China, Chengdu, 611731, China\\
$^2$ Institute of Electronic and Information Engineering of UESTC, Dongguan, 523808, China\\
$^3$Department of Computer Science and Engineering, Lehigh Univerisity, Bethlehem, USA\\
\emails
zongmohuang@gmail.com,\{yazhou.ren,puxiaor\}@uestc.edu.cn,lih319@lehigh.edu
}

\begin{document}

\maketitle

\begin{abstract}
With the representation learning capability of the deep learning models, deep embedded multi-view clustering (MVC) achieves impressive performance in many scenarios and has become increasingly popular in recent years.
Although great progress has been made in this field, most existing methods merely focus on learning the latent representations and ignore that learning the latent graph of nodes also provides available information for the clustering task.
To address this issue, in this paper we propose Deep Embedded Multi-view Clustering via Jointly Learning Latent Representations and Graphs (DMVCJ), which utilizes the latent graphs to promote the performance of deep embedded MVC models from two aspects.
Firstly, by learning the latent graphs and feature representations jointly, the graph convolution network (GCN) technique becomes available for our model. 
With the capability of GCN in exploiting the information from both graphs and features, the clustering performance of our model is significantly promoted.
Secondly, based on the adjacency relations of nodes shown in the latent graphs, we design a sample-weighting strategy to alleviate the noisy issue, and further improve the effectiveness and robustness of the model. 
Experimental results on different types of real-world multi-view datasets demonstrate the effectiveness of DMVCJ.
\end{abstract}

\section{Introduction}
As the fundamental task in the unsupervised learning field, clustering has been extensively studied for many years and numerous classical clustering algorithms have been developed, such as kmeans \cite{MacQueen:some}, spectral clustering \cite{spectralclustering}, density-based clustering \cite{ester1996density} and hierarchical clustering \cite{Jain:clustering} have been developed.
However, conventional clustering methods can only handle single-view data,
while in the real-world most objects can be described from multiple aspects or views.
 
To make full use of the complementary information from multiple views and obtain better clustering results, in the past decade, a series of multi-view clustering (MVC) algorithms have been proposed \cite{Kumar2011ACA,Tzortzis2012KernelBasedWM,Cai2013MultiViewKC,XU201625,zhang2017multi,DEMVC}. 
Among these methods, the recently emerged deep embedded MVC models \cite{DAMC,DEMVC,o2mgc} achieve superior clustering performance over many state-of-the-art methods and attract increasing attentions.


Although great progress has been made, few deep embedded MVC methods make use of the graph structure information in their model, which prevent them from achieving better results.
Meanwhile, despite some latest graph convolution network (GCN) \cite{GCN} based models \cite{MAGCN,o2mgc} exploit information from both graph structure and node features, their application scenarios are extremely limited as these methods require explicit graph data as input.
 
To address the above issues, in this paper, we propose Deep Embedded Multi-view Clustering via Jointly Learning Latent Representations and Graphs (DMVCJ).
Concretely, we firstly apply a series of auto-encoders \cite{AE} to initialize the latent representations of nodes in each view.
Then, in each epoch of the finetuning stage, we dynamically construct adjacent graph on latent representations for each view via the kNN algorithm \cite{knn}.
After that, we perform graph convolution on the embedded node features with the obtained graphs to enhance the latent representations before clustering.
Specifically, in the beginning of every $T$ iterations, we generate self-supervised pseudo-labels for the samples based on latent representations refined by GCN.
By taking advantage of graph information, the distribution of the refined latent representations reflect the cluster structure of data more accurately, thus the pseudo-labels derived by them are useful in promoting the representation learning of model.
As a return, the promotion of the latent representation capability of embedding encoders also contributes to learning more precise latent graph.
Therefore, DMVCJ can achieve better clustering performance by jointly learning the latent representations and graphs in a mutually beneficial manner.

Moreover, based on the adjacent graphs of each view, we further design a simple yet effective sample-weighting approach to alleviate the noisy issue.
In the GCN module, it aggregates the features of neighboring nodes to obtain the representation of target node. 
It is not hard to aware that the nodes with higher in-degree values generally locate near the center of clusters, while the nodes with very small in-degree values could be a noisy point.
Based on this theory, during the finetuning process, we assign higher weights to the nodes with higher in-degree values and vice versa.
By this way, our method alleviates the impact from noises and thus promotes the performance and robustness of model.
 
In summary, the main contributions of the paper include:
\begin{itemize}
\item A novel deep embedded multi-view clustering network which jointly learns the latent graph structure and feature representation is proposed for the multi-view data where explicit graphs are unobserved.
\item A simple yet effective sample-weighting strategy based on the in-degrees in the latent adjacent graph is designed to alleviate impact from noisy samples.
\item Experiments on different types of real-world multi-view datasets demonstrate the effectiveness of our algorithm. 
\end{itemize}

\section{Related Work}
With the capability in exploiting the complementary information of different views, multi-view clustering becomes a hot research topic in the past decade and numerous MVC models \cite{DAMC,DEMVC,SDMVC,AAAIVAE,ICCVVAE} have been proposed.
Among these models, the deep embedded MVC models achieve impressive results in many scenarios by learning representative latent features through elaborate deep networks. 
For instance, \cite{DAMC} introduces the adversarial network modules to help the model learn more discriminating embedded features and thus promotes the clustering results.
In DEMVC \cite{DEMVC}, the authors promote the representation capability of model by training multiple deep neural networks corresponding to different views collaboratively.
In another work \cite{SDMVC}, they propose to avoid the impact of certain views with unclear clustering structures by aligning the embedded representation of different views.
Moreover, some deep embedded MVC models \cite{ICCVVAE,AAAIVAE} adopt variation auto-encoder (VAE) \cite{VAE} to extract the disentangled latent features of each sample, with the interpretability of the features learned by VAE, these models have much wider application potentials.
 
Though various attempts were made to improve the effectiveness of the deep embedded models based on the above research, most of them ignore the fact that the graph structure of nodes also provides available information of the clustering task, which prevent them from achieving better results.
 Inspiringly, the emergence of graph convolution networks made some difference, with GCN's capability in utilizing information from both graph structure and node features.
 There are some studies that have been proposed in the literature for applying GCN on the deep embedded MVC models. 
O2MGC \cite{o2mgc} firstly employs GCN techniques to the multi-view clustering task, in which the representation of the most informative view is shared with others.
In MAGCN \cite{MAGCN}, the authors reconstruct both graph structure and data features as well as maintains the consistency of geometric relationship and clustering distribution in different views.
 
Although some progress has been made by the GCN-based deep embedded models, they require explicit graph data as input. 
In most real-world applications, multi-view data merely contain nodes features, thus the application scenario of these models is extremely limited.
Moreover, these models perform graph convolution on the features of original data and the first few layers of network.
However, those features are generally not representative enough, which may leave negative influence on the representation learning of models and thus obtain suboptimal clustering results.
 
Unlike the above methods, the proposed DMVCJ not only jointly learns both latent representations and the latent graphs in a mutually beneficial manner, but is also applicable to to the multi-view data without explicit graphs.

\section{Proposed Method}
\textbf{Problem Statement.} Given a multi-view dataset $X={\{X^v\}}^m_{v=1}$ with $m$ views, where $X^v = \{x^v_1;x^v_2;\ldots;x^v_n\} \in R^{n \times d^v}$, $d^v$ is the dimension of the feature vector in the $v$-th view and $n$ is the instance number.
Our target is to partition $n$ instances into $k$ clusters. 
Specifically, we aim to better clustering results by jointly learning the graph structures and feature representations among different views.
 
\subsection{Model}
\begin{figure*}[!t] 
\begin{center}
\includegraphics[width=1.0\linewidth]{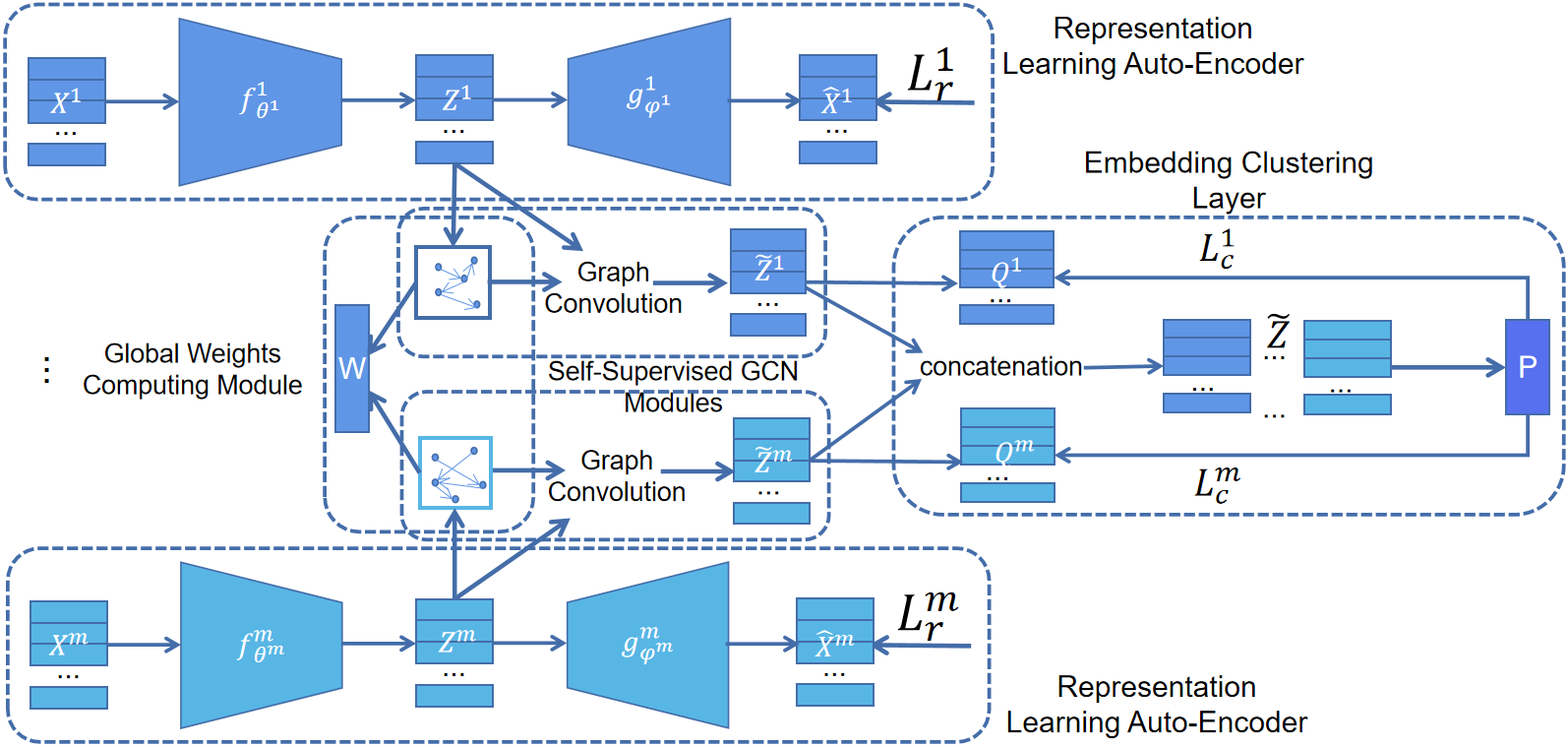}
\end{center}
   \caption{The framework of DMVCJ.
   For the $v$-th view, the representation learning auto-encoder learns the latent representations $Z^v$ of the original data $X^v$.
   After that, the self-supervised GCN module generates the adjacent graphs $A^v$ of the embedded nodes, and performs graph convolution to obtain the refined representations $\Tilde{Z}^v$. 
    Meanwhile, the adjacent graphs of different views are also sent to the global weights computing module to detect the global importance of the samples $W$.
    Finally, in the embedding clustering layer, the $\Tilde{Z}^v$ from different views are concatenated as $\Tilde{Z}$, which is fed to kmeans algorithm weighed by $W$ to obtain the clustering result.
   }
\label{model} 
\end{figure*}
Figure \ref{model} provides an overview of the proposed DMVCJ, which mainly consists of four modules: 1) representation learning auto-encoder, 2) self-supervised GCN module, 3) global weights computing module, and 4) embedding clustering layer.
We describe each module in detail as follows.
\subsubsection{Representation Learning Auto-Encoder} 
Following the existing deep embedded MVC models, in DMVCJ, the deep auto-encoder is applied to learn the embedded features of nodes in different views.
Specifically, we use $f^v_{\theta^v}$ and $g^v_{\phi^v}$ to represent the encoder and decoder of the $v$-th view where $\theta^v$ and $\phi^v$ are learnable parameters.
Then the embedding process can be written as:
\begin{equation}
\label{embedding}
    z^v_i=f^v_{\theta_v}(x^v_i),
\end{equation}
 where $z^v_i \in R^{d_v}$ denotes the latent representation of $x^v_i$ in the $d_v$-dimensional space.
After that, the decoder $g^v_{\phi^v}$ decodes $z^v_i$ to reconstruct the original data as $\hat{x}^v_i$:
\begin{equation}
    \hat{x}^v_i=g^v_{\phi_v}(z^v_i).
\end{equation}
The reconstruction loss of the $v$-th view is defined as follows:
\begin{equation}
    L^v_r=\sum\limits^{n}_{i=1}{{||x^v_i-g^v_{\phi^v}(f^v_{\theta^v}(x^v_i))||}^2_2}.
\label{lr}
\end{equation}
By optimizing the reconstruction loss, the auto-encoders learn the latent representation of each sample.
 
\subsubsection{Self-Supervised GCN Module} 
To enhance the model's representation capability via graph information, we introduce a two-layer self-supervised GCN module for each view which learns the graph structure of nodes in a self-supervised manner.
 
Let $Z^v = \{z^v_1;z^v_2;\ldots;z^v_n\} \in R^{n \times d_v}$ denote the latent representations of all the samples in the $v$-th view. 
The self-supervised GCN module first constructes the adjacent matrix $A^v \in R^{n \times n}$ using the kNN algorithm according to the distances of the embedded representations.
Following the latest research about graph convolution \cite{S2GG}, this GCN module merely performs linear mapping, thus the refined embedded representation $\Tilde{z}^v_i$ is computed by:
\begin{equation}
    \Tilde{z}^v_i={(\Tilde{D}^{-\frac{1}{2}}\Tilde{A}\Tilde{D}^{-\frac{1}{2}})}^2{z}^v_i,
\end{equation}
where $\Tilde{A} = I_n+A$ and $\Tilde{D}_{ii}=\sum_{j}\Tilde{A}_{ij}$.
Through the self-supervised GCN module, our model utilizes graph information in the embedding process and thus helps mine the cluster structures more accurately.
 
In addition, with the existence of self-supervised GCN module, our method has two advantages compared with the existing GCN-based MVC models.
Firstly, unlike existing GCN-based MVC models, our method does not require explicit graph data as input, so that it has much wider application scenarios.
Secondly, instead of performing graph convolution on the original data and the first few layers of networks where lots of corrupt and redundant features exist, our method performs graph convolution on the more representative latent features and thus makes better use of the graph information.

\subsubsection{Global Weights Computing Module}
To alleviate the impact from noisy samples, a global sample weighting vector $W=[w_1, w_2,\dots,w_n]$ derived by the adjacent graphs from multiple views is also introduced to our model.
Specifically, for the $i$-th node in the $v$-th view, its in-degree ${\eta}^v_i$ in the refined adjacent matrix $\Tilde{A}^v$ is:
\begin{equation}
    {\eta}^v_i=\sum\limits_{i}\Tilde{A}^v_{ij}.
\end{equation}
As in the self-supervised GCN module, $\Tilde{A}^v$ is derived by the kNN algorithms, for the $i$-th node, the value of ${\eta}^v_i$ means the number of nodes (including itself) that regard it as a neighbor in the $v$-th view.
It is not hard to aware that the nodes with higher ${\eta}^v_i$ generally locate near the center of clusters, while the nodes with very small in-degree values are probably noisy points.
Extending this theory to multiple views, then the value of $\eta_i=\sum_{v=1}^{m}{\eta}^v_i$ reflects global importance of the $i$-the sample.
Then the global weight of the $i$-th instance $w_i$ is computed by a simple linear function:
\begin{equation}
    w_i=min(\eta_i/\lambda,1) \in (0,1],
\label{wi}
\end{equation}
where $\lambda$ is set to the median value of $\eta=[\eta(1),\eta(2),\dots,\eta(n)]$ confirming that at least half of the samples are treated as normal samples as well as all the samples can take part in the training.
 By assigning larger weights to more important nodes and vice versa during the training process, it can alleviate the negative influence of noises and obtain better performance and robustness.
 
\subsubsection{Embedding Clustering Layer}
After obtaining the refined embedded representation $\Tilde{Z}^v = \{\Tilde{z}^v_1;\Tilde{z}^v_2;\ldots;\Tilde{z}^v_n\} \in R^{n \times d_v}$, an embedding clustering layer $c^v_{\mu^{v}}$ is applied to compute the cluster assignments of samples in the $v$-th view, where $\mu^{v}$ denotes the learnable cluster centroid. 
Specifically, based on the widely used Student’s $t$-distribution \cite{tsne} in deep embedded clustering models, in the $v$-th view, the probability that the $i$-th example belongs to the $j$-th cluster is:
\begin{equation}
    q^v_{ij}=c^v_{\mu^{v}}(\Tilde{z}^v_i)=\frac{{(1+{||\Tilde{z}^v_i-\mu^v_j||}^2)}^{-1}}{\sum\limits_{j}{(1+{||\Tilde{z}^v_i-\mu^v_j||}^2)}^{-1}}.
\end{equation}
 
Let $Q^v=\{q^v_1;q^v_2;\ldots;q^v_n\}\in R^{n \times k}$ represent the cluster assignments of all the samples in the $v$-th.
In this module, we adopt the self-supervised training idea developed in \cite{SDMVC} i.e, promoting the representation capability of the embedded features in each view by minimizing the difference between the single-view cluster assignment $Q^v$ and the global pseudo-label $P$.
 
Concretely, $P$ is obtained by the following procedures:
Firstly, for each sample, concatenating its embedded representations in all the views:
\begin{equation}
    \Tilde{z}_i = [\Tilde{z}^1_i ,\Tilde{z}^2_i ,\ldots,\Tilde{z}^m_i] \in R^{\sum_{v=1}^{m}d_v}.
\end{equation}
Then, we apply the weighted-kmeans to generate the global cluster centroids $c_j$:
\begin{equation}
    \min\limits_{c_1 ,c_2 ,\ldots,c_k }\sum\limits_{i=1}^{n}\sum\limits_{j=1}^{k}w_i{||\Tilde{z}^i-c_j||}^2.
\end{equation}
Computing the soft assignment $s_{ij}$ between each global embeddings and each cluster centroid with Student’s $t$-distribution:
\begin{equation}
        s_{ij}=\frac{{(1+{||\Tilde{z}_i-c_j||}^2)}^{-1}}{\sum_{j}{(1+{||\Tilde{z}_i-c_j||}^2)}^{-1}}.
\end{equation}
Finally, the global pseudo-label $P$ is computed by:
\begin{equation}
    p_{ij}=\frac{(s_{ij}^2/\sum_{i}s_{ij})}{\sum_{j}(s_{ij}^2/\sum_{i}s_{ij})},
\label{P}
\end{equation}
where $P_{ij}$ denote the probability that the $i$-th example belongs to the $j$-th cluster.
 
After obtaining $P$, we define the clustering loss $L^v_c$ for each view as the Kullback-Leibler divergence (DKL) between the pseudo-label $P$ and $Q^v$:
\begin{equation}
    L^v_c=D_{KL}(P||Q^v)=\sum\limits_{i=1}^{n}w_i\sum\limits_{j=1}^{k}p_{ij}log\frac{p_{ij}}{q_v^{ij}} 
\label{lc}
\end{equation}
 
By optimizing Eq. (\ref{lc}), the representation capability of the embedding auto-encoders of each view will be promoted with the information from multiple views, and thus enhancing multi-view clustering performance.
 
Using the sample weights Eq. (\ref{wi}) to refine the reconstruction loss Eq. (\ref{lr}), and integrating with clustering loss Eq. (\ref{lc}), the loss function for DMVCJ in each view is:
\begin{equation}
    L^v=\sum\limits^{n}_{i=1}w_i({{||x^v_i-g^v_{\phi^v}(f^v_{\theta^v}(x^v_i))||}^2_2}+\gamma\sum\limits_{j=1}^{k}p_{ij}log\frac{p_{ij}}{q_v^{ij}}),
\label{tl}
\end{equation}
where $\gamma$ is a trend-off coefficient.

\subsection{Optimization}
The optimization of DMVCJ consist of two steps, i.e., initialization and finetuning.
In initialization, we firstly train auto-encoders $f^v_{\theta^v}$ and $f^v_{\phi^v}$ of each view by optimizing the reconstruction loss in Eq. (\ref{lr}).
After that, the initial $c^v_{\mu^v}$ for each view is obtained by kmeans algorithm. 
 
Then, in the beginning of every $T$ iterations in the finetuning stage, the global sample weights $w_i$ and pseudo-labels $P$ are obtained by Eq. (\ref{wi}) and Eq. (\ref{P}) respectively.
With the capability of GCN in exploiting both graph and feature information, the self-supervised pseudo-labels reflect the cluster distribution of data more accurately and thus improve the model's representation learning performance when minimizing the value of Eq. (\ref{tl}) during the training. 
Besides, by introducing sample-weights $w$ to alleviate the noisy issue, the effectiveness and robustness of the model are further improved.
Specifically, our method keep running until the maximum iteration number $T_{max}$ is reached.
 
After the training process terminates, we compute the pseudo-label $P$ once again and the final clustering assignment $y_i$ for the $i$-th sample is obtained by:
\begin{equation}
    y_i=\mathop{\arg\max}_{j}(P_{ij}).
\label{Y}
\end{equation}
 
The specific hyper-parameter settings of DMVCJ are illustrated in Section \ref{settings}, and the entire workflow of DMVCJ is summarized in Algorithm \ref{alg:algorithm}. 

\begin{algorithm}[!t]
 \renewcommand{\algorithmicrequire}{\textbf{Input:}} 
 \renewcommand{\algorithmicensure}{\textbf{Output:}} 
\caption{The DMVCJ model.}
 \label{alg:algorithm} 
\begin{algorithmic}[1]
\REQUIRE ~~
Data set $X^v$, $v=1,2,\ldots,m$; Cluster number $k$; 
\ENSURE 
Cluster assignments $Y=\{y_1,y_2,\dots,y_n\}$.
\STATE Pretrain the auto-encoder separately in each view by optimizing Eq. (\ref{lr}).
\STATE Initialize the cluster centroids by kmeans algorithms.
\WHILE{Not Reach the maximum iteration $T_{max}$}
\STATE Update the global sample weights $W$ and pseudo-labels $P$ according to Eq. (\ref{wi}) and Eq. (\ref{P}).
\REPEAT
\STATE Finetuning the whole network by optimizing Eq. (\ref{tl}).
\UNTIL{The iteration time is divisible by $T$.}
\ENDWHILE 
\STATE Get the final pseudo-labels $P$ according to Eq. (\ref{P}).
\STATE Compute $y_i$ for each sample according to Eq. (\ref{Y}).
\end{algorithmic}
\end{algorithm}


\section{Experiments}

\begin{table*}[!tbh]
\small
\setlength\tabcolsep{3pt}
\renewcommand\arraystretch{1.6}
\begin{tabular}{c|c|c|c|c|c|c|c|c|c}
\hline
Dataset        & \multicolumn{3}{c|}{BDGP}                      & \multicolumn{3}{c|}{Handwritten Numerals}                     & \multicolumn{3}{c}{Reuters}                \\ \hline
Info        & \multicolumn{3}{c|}{1 visual view and 1 textual view,} & \multicolumn{3}{c|}{6 visual views,}        & \multicolumn{3}{c}{5 textual views,}       \\
        & \multicolumn{3}{c|}{k=5, n=2500.}              & \multicolumn{3}{c|}{k=10, n=2000.}          & \multicolumn{3}{c}{k=6, n=1200.}           \\ \hline
Methods & ACC(\%)           & NMI(\%)             & ARI(\%)             & ACC(\%)            & NMI(\%)            & ARI(\%)            & ACC(\%)            & NMI(\%)            & ARI(\%)            \\ \hline
KM       &57.68(2.93)   &47.35(2.43)   &19.38(3.90)   &75.45(5.00)  &78.58(4.14)  &66.72(4.32)  &29.12(6.33)  &13.53(8.87)  &6.80(6.33) \\
SC       &59.98(8.17)               &50.98(5.57)               &26.20(5.14)               &77.69(0.08)              &86.91(0.15)              &75.26(0.17)              &17.89(0.11)              &2.71(0.24)              &0.05(0.01)              \\
IDEC       &91.28(7.55)               &85.64(6.95)               &81.98(8.82)               &84.13(8.65)              &84.61(3.97)              &78.37(8.03)              &45.98(2.78)              &25.17(2.71)              &18.02(2.22)              \\ \hline
MVKKM  &42.02(3.02)               &27.33(1.78)               &12.69(1.53)               &67.21(3.09)              &67.70(0.28)              &55.76(0.79)              &24.81(5.82)              &11.67(6.79)              &3.76(3.57)              \\ 
MLAN       &47.32(0.00)               &31.30(0.00)               &24.29(0.00)               &97.35(0.00)              &94.00(0.00)              &94.17(0.00)              &21.50(0.00)              &15.04(0.00)              &1.49(0.00)              \\
AMVCD  &56.87(5.41)               &43.58(5.72)               &22.71(4.29)               &79.66(9.32)              &84.90(3.58)              &75.99(8.69)              &27.38(2.73)              &10.51(2.58)              &4.29(1.65)              \\ 
GMC  &59.12(0.00)               &62.61(0.00)               &43.13(0.00)               &88.20(0.00)              &90.73(0.00)              &85.40(0.00)              &19.75(0.00)              &12.95(0.00)              &1.29(0.00)               \\
SAMVC  &51.31(7.48)               &45.15(6.49)               &19.60(5.96)               &76.37(7.36)              &84.41(2.50)              &73.87(6.25)              &18.83(1.92)              &4.58(3.40)              &0.32(0.62)              \\ 
DEMVC  &92.78(1.55)               &83.31(3.35)               &82.64(3.89)               &67.69(6.15)              &70.61(2.99)              &58.86(4.92)              &46.71(0.85)              &25.31(1.43)              &20.41(1.06)              \\ 
SDMVC  &97.89(0.52)               &93.41(1.24)               &94.85(1.20)               &97.18(0.51)              &94.44(0.52)              &93.93(0.97)              &47.07(0.79)              &27.12(1.20)              &21.22(0.99)              \\ 
DMVCJ  &\textbf{98.57(0.24)}               &\textbf{95.42(0.54)}               &\textbf{96.54(0.47)}               &\textbf{98.03(0.26)}              &\textbf{95.46(0.47)}              &\textbf{95.65(0.59)}              &\textbf{54.77(2.82)}              &\textbf{33.52(2.68)}              &\textbf{26.20(2.00)}              \\ \hline
\end{tabular}
\caption{Clustering results of compared methods on three multi-view datasets.}
\label{TabRes}
\end{table*}

\subsection{Experimental Setup}
\textbf{Datasets}
We use the following three multi-view datasets in our study, which are publicly available:

\textbf{BDGP} \cite{BDGP} contains 2,500 images about 5 different categories of drosophila embryos. 
Each image has two views: visual features (1750 dimensions) and textual features (79 dimensions).

\textbf{Handwritten Numerals}\footnote{https://archive.ics.uci.edu/ml/datasets.php} comes from UCI machine learning repository, which contains 2000 points in 10 classes corresponded to numerals (0-9). 
Each instance is constituted by the six visual views i.e., 216 profile correlations, 76 Fourier coefficients of the character shapes, 64 Karhunen-Love coefficients, 6 morphological features, 240 pixel averages in 2 $\times$ 3 windows, and 47 Zernike moments.
 
\textbf{Reuters}\footnote{http://lig-membres.imag.fr/grimal/data.html} selects 1200 articles from 6 categories (C15, CCAT, E21, ECAT, GCAT and M11), each providing 200 articles. 
Each document is written in five different languages (English, French, German, Italian, and Spanish), corresponding to five textual views.
 
\paragraph{Comparing Methods} 
To demonstrate the effectiveness of the proposed DMVCJ, we compare it with seven existing state-of-the-art multi-view clustering methods:
\begin{itemize}
\item MVKKM: Kernel-Based Weighted Multi-View Clustering \cite{Tzortzis2012KernelBasedWM}.
\item MLAN: Multi-View Clustering and Semi-Supervised Classification with Adaptive Neighbours \cite{MLAN}.
\item AMVCD: Auto-Weighted Multi-View Clustering via Deep Matrix Decomposition \cite{HSDDMVC}.
\item GMC: Graph-Based Multi-View Clustering \cite{GMC}.
\item SAMVC: Self-Paced and Auto-Weighted Multi-View Clustering \cite{REN2020248}.
\item DEMVC: Deep Embedded Multi-View Clustering with Collaborative Training \cite{DEMVC}.
\item SDMVC: Self-supervised Discriminative Feature Learning for
Deep Multi-View Clustering \cite{SDMVC}.
\end{itemize}
 
To make a comprehensive comparison, we also employ some single-view methods, i.e., KMeans (KM) \cite{MacQueen:some}, Spectral Clustering (SC) \cite{spectralclustering}, and Improved Deep Embedded Clustering (IDEC) \cite{IDEC} by concatenating features from all the views.

\begin{figure*}[!tbh]
\centering
\begin{minipage}{0.33\linewidth}
\centering
\includegraphics[width=2.3in, height = 1.78in]{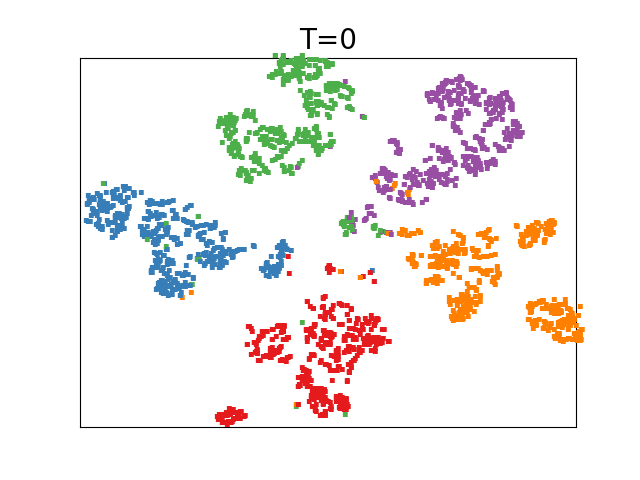}
\end{minipage}%
\begin{minipage}{0.33\linewidth}
\centering
\includegraphics[width=2.3in, height = 1.78in]{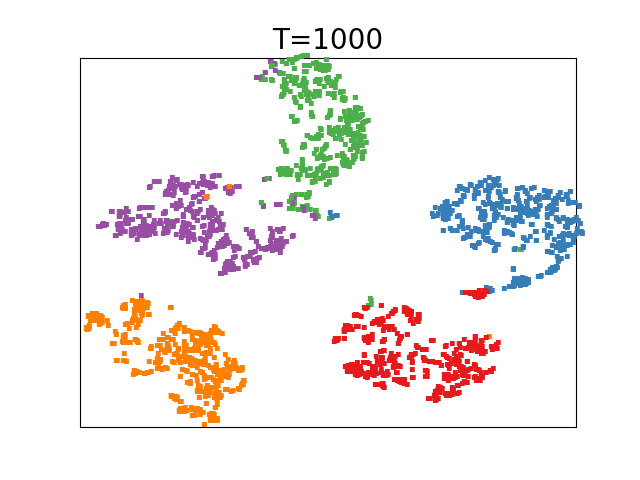}
\end{minipage}
\begin{minipage}{0.33\linewidth}
\centering
\includegraphics[width=2.3in, height = 1.78in]{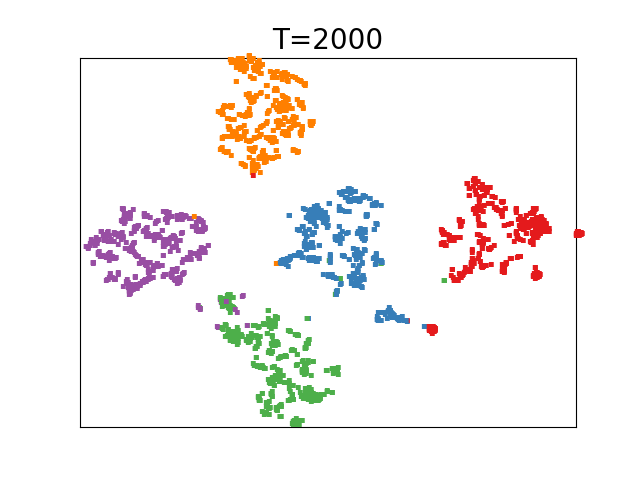}
\end{minipage}%
\caption{Visualization of the refined latent representations of BDGP dataset during the training.}
\label{bdgpt}
\end{figure*}

\begin{figure*}[!t]
\centering
\begin{minipage}{0.33\linewidth}
\centering
\includegraphics[width=2.3in, height = 1.78in]{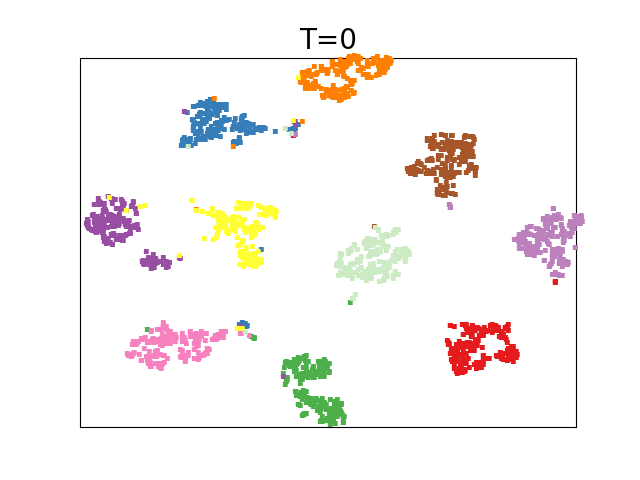}
\end{minipage}%
\begin{minipage}{0.33\linewidth}
\centering
\includegraphics[width=2.3in, height = 1.78in]{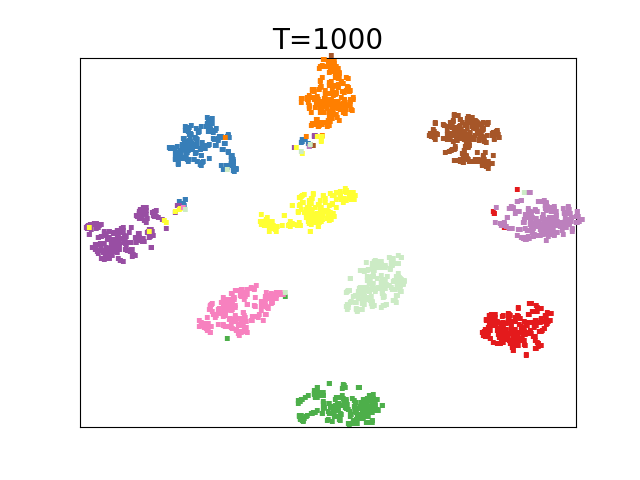}
\end{minipage}
\begin{minipage}{0.33\linewidth}
\centering
\includegraphics[width=2.3in, height = 1.78in]{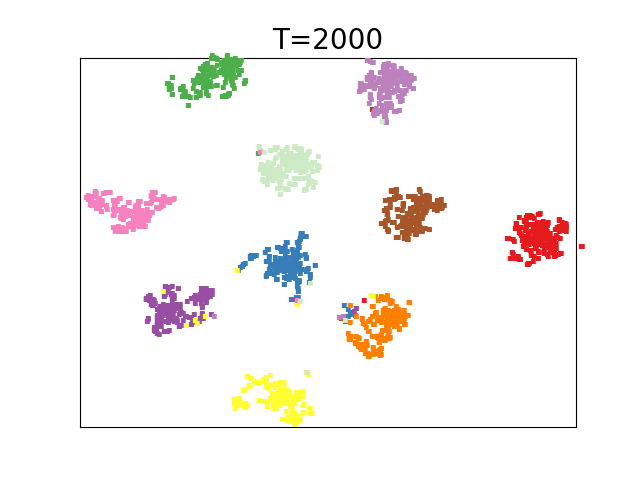}
\end{minipage}%
\caption{Visualization of the refined latent representations of Handwritten Numerals dataset during the training.}
\label{hwt}
\end{figure*}
 
\paragraph{Implementation Details}\label{settings} 
For all the three datasets, we use the same fully connected auto-encoder structure as \cite{IDEC}, i.e., for each view, the structure of encoder is: Input($d^v$) - Fc500 - Fc500 - Fc2000 - Fc10($d_v$).
Decoders are symmetric with the encoders of corresponding views. 
ReLU \cite{relu} is used as the activation function and Adam \cite{adam} (default learning rate is 0.001) is chosen as the optimizer. 
All the auto-encoders are pre-trained for 2000 epochs. 
The trade-off coefficient $\gamma$ is set to 0.1 and the value of $k$ in the kNN algorithms is set to 10. 
As our method needs to build the graph on all the nodes, the batch size is equal to the instance number $n$.
The pseudo-label $P$ and the global weights $W$ are updated in the beginning of every $T=1000$ epochs, and the whole training process terminates when the epoch number reaches $T_{max}=5000$.
 
For the comparing methods, we directly use the source codes provided by the authors and the parameter settings therein.
Specifically, since SDMVC cannot naturally terminate on the Handwritten Numerals dataset, we stop it manually after a large iteration number 30000.
 
\paragraph{Evaluation Measures} 
To evaluate the clustering results, we employ the widely used metrics accuracy (ACC), normalized mutual information (NMI) and adjusted rand index (ARI) for quantitative evaluations, where higher values indicate better clustering results.
We report the average results and standard deviations of ten independent runs of each method on three datasets.

\subsection{Clustering Results}
In this subsection, we demonstrate the effectiveness of the proposed DMVCJ through the comparison with the baselines and the embedding visualization during training.


\textbf{Comparison with Baselines}
Table \ref{TabRes} shows the clustering results of the baseline methods compared with DMVCJ.
In each column, the best results are highlighted in boldface.
From the results, we can observe that DMVCJ achieves superior performance on all three metrics across three multi-view datasets with different view compositions.
This is mainly because DMVCJ jointly learns both latent representations and graphs based on a self-supervised GCN module.
With the information of the latent graphs, our method enhances the latent representations with the capability of GCN module, which helps our model to mine the clusters more accurately.
Moreover, based on the in-degrees values within the latent adjacent graph, our method assigns smaller weights to the noisy samples during the training, thus further promotes the effectiveness as well as the robustness of model.  

\textbf{Visualization of Learning Process}
To visualize the effectiveness of our method, we use the $t$-SNE algorithm \cite{tsne} to reduce the dimension of the extracted feature vectors to 2D and demonstrate the separability/non-separability of the data at different iterations $T= 0, 1000, 2000$.
Figures \ref{bdgpt} and \ref{hwt} show the visualization results on BDGP and Handwritten Numerals datasets, where different colors denote the labels of corresponded nodes.
It can be seen that our method achieves remarkable clustering performance even in the beginning of the finetuning stage, and as the training forwards, the cluster structures become increasingly clear, which demonstrates the effectiveness of the proposed DMVCJ.

\section{Conclusion and Future Work}
In this paper, we proposed Deep Embedded Multi-view Clustering via Jointly Learning Latent Representations and Graphs (DMVCJ) for multi-view clustering, which aims to promote the clustering performance of deep embedded MVC models with the latent graph information.
Specifically, by introducing a self-supervised GCN module, DMVCJ jointly learns both latent graph structures and feature representations in a mutual benefit manner and significantly promotes the clustering performance.
In addition, we designed a global sample-weighting strategy based on the in-degrees of nodes in the adjacent graphs to alleviate the impact from noisy samples. The experiments on three types of multi-view data sets demonstrated the effectiveness of the proposed DMVCJ.
Nevertheless, the quality of each view is not considered in this work, it is an interesting future work to consider learning weights for different views to further improve our model.

\newpage

\bibliographystyle{named}
\bibliography{output}
\end{document}